\documentclass{article}
\usepackage{spconf,amsmath,amssymb,graphicx,booktabs,multirow,url}
\usepackage{tipa}
\usepackage{textcomp}
\usepackage{newunicodechar}
\usepackage[style=ieee, backend=biber]{biblatex}
\usepackage[hidelinks]{hyperref}

\usepackage{array}

\newunicodechar{ɖ}{\textipa{\:d}}
\newunicodechar{Ɖ}{\textipa{\:D}}
\newunicodechar{ɛ}{\textipa{E}}
\newunicodechar{ƒ}{\textflorin}
\newunicodechar{ɣ}{\textipa{\textgamma}}
\newunicodechar{ŋ}{\textipa{N}}
\newunicodechar{ɔ}{\textipa{O}}
\newunicodechar{ʋ}{\textipa{V}}
\newcommand{\textltailgamma}{\textipa{\;G}}

\title{Nufo: A Benchmark for Ewe Automatic Speech Recognition}

\name{Christ Kekeli Keli$^{\star}$, Steve Sanogo$^{\star}$, Patrice Sebastiano$^{\star}$, Hamad Aldenawi$^{\star}$, Ryan Whetten$^{\dagger}$}
\address{$^{\star}$Universit\'e d'Avignon, LIA, France \\
$^{\dagger}$Brigham Young University–Hawaii, Laie, Hawaii}

\begin{document}
\maketitle

\begin{abstract}
Automatic speech recognition (ASR) remains challenging for low-resource languages, where the scarcity of transcribed speech limits the development of accurate systems. This challenge is particularly acute for African languages such as Ewe, which has millions of speakers but remains underrepresented in publicly available ASR resources and benchmarks. In this work, we consolidate, normalize, and partition existing Ewe speech resources into a standardized 174-hour Ewe ASR benchmark, the largest to date to our knowledge. We fine-tune and evaluate pretrained speech models, including Whisper and AfriHuBERT, to establish strong baselines on our benchmark. Whisper-medium achieves a word error rate (WER) of 37.98\%, while AfriHuBERT achieves 39.70\% on the test set. We further provide a qualitative analysis of common recognition errors. We release the benchmark under the name Nufo, derived from the Ewe word for ``talk", accompanied by models and evaluation resources, to facilitate future research on Ewe and other underrepresented African languages.
\end{abstract}


\section{Introduction}
\label{sec:intro}

Automatic speech recognition (ASR) has reached high accuracy in
high-resource languages, such as English, with word error rates (WERs) under 10\%~\cite{pmlr-v202-radford23a}. 
However, African languages remain substantially under-represented in corpora, benchmarks, and pretrained models resulting in WERs anywhere from 30\% to 60\%~\cite{olufemi2026waxal, alabi25_interspeech}. 
The bottleneck is rarely an absence of speakers; it is the lack of large, diverse, openly accessible speech data with reliable transcriptions, which makes it difficult to adapt modern end-to-end architectures that were pretrained predominantly on high-resource languages. 
Ewe is a representative case. 

Ewe (Eʋegbe) is a Niger-Congo language of the Gbe group spoken primarily in Ghana and Togo, with speakers also in Benin, and has approximately five million native speakers. 
Ewe has received far less attention in speech technology than high-resource languages. 
Additionally, Ewe is tonal, meaning that pitch variation can carry linguistic distinctions, and its orthography contains characters absent from the standard English alphabet (i.e. ɖ, ɛ, ƒ, ɣ, ŋ, ɔ, ʋ). These properties can further complicate speech and text processing.

Recent initiatives have begun to relieve the data shortage. 
BibleTTS~\cite{meyer22c_interspeech} contributes aligned Ewe speech--text pairs that, although collected for text-to-speech, could also be utilized ASR. 
Common Voice - Mina~\cite{ardila-etal-2020-common} provides transcribed Mina speech. Mina (or Gen) is a Gbe language variety closely related to Ewe in speech an orthography. 
Furthermore, the WAXAL project~\cite{diack2026waxal} supplies a large-scale multilingual African corpus that includes Ewe. 
Despite these efforts, the amount of transcribed Ewe remains minimal compared to high-resource languages, with the largest Ewe speech datasets containing around 100 hours of transcribed data.

Alongside the lack of labeled data, these resources were developed with different collection procedures, recording conditions, speaker distributions, transcription conventions, preprocessing, and partitions. Due to collection procedures, these resources cannot be treated as completely independent sources; recordings or utterances may be shared across corpora. As a result, pooling them without a prior audit risks leakage between training and evaluation data and artificially optimistic scores, particularly in a low-resource setting where each recording carries substantial weight. Additionally, the lack of an accessibility standard benchmark for Ewe ASR, leads to challenges when it comes to the reproduction and direct comparison of methodologies and model.

To address these challenges in Ewe ASR, we make the following contributions.

\begin{itemize}
\item First, we introduce Nufo, an Ewe ASR dataset, with approximately 174 hours of audio and standardized training, validation, and test splits.\footnote{\url{https://huggingface.co/datasets/medbridge-ai/Nufo}}.
\item Second, we release fine-tuned Ewe ASR models based on Whisper and AfriHuBERT and compare their performance.
\item Third, we provide a qualitative analysis of Ewe ASR performance, identifying common error patterns and highlighting challenges for Ewe speech recognition.
\end{itemize}

\section{Related Work}
\label{sec:related}

Ewe was largely absent from early multilingual speech resources such as GlobalPhone~\cite{schultz02_icslp} and Multilingual LibriSpeech~\cite{Pratap2020MLSAL}. Leaving early work on low-resource African language dependent on small, domain-restricted corpora such as religious speech from the BibleTTS~\cite{meyer22c_interspeech} dataset, whose limited speaker diversity makes them unsuitable as general benchmarks for speech recognition. 

Recently, researchers held a crowdsourcing effort for Ewe ASR yielding 203,336 validated recordings, approximately 1,130~h from 1,937 speakers of which about 107~hours are transcribed~\cite{wiafe-etal-2025-building}. 
These were collected using culturally relevant images as prompts rather than sentence readings, as crowdsourcing by asking people to read out loud sentences can be problematic for speakers with limited literacy in Ewe. Fine-tuning multilingual Whisper, they report 37.12\% WER and 12.43\% CER, and 37.17\% WER and 12.50\% CER with the language set to Yoruba and Shona, respectively. 

In the WAXAL project, researchers assemble a large African corpus and a dedicated ASR benchmark covering 24 languages including over 1,000 hours of Ewe, in which 99.8 hours are transcribed~\cite{diack2026waxal}. The subsequent work, WAXAL-Net~\cite{olufemi2026waxal} establishes Ewe ASR baselines (along with baselines for other languages) by fine-tuning multilingual pretrained speech models of various sizes. For Ewe, they report WERs ranging from 31.3\% to over 88.7\% depending on the pretrained model employed. One highlight from WAXAL-NET is that researchers found that relatively small specialized model (from 39M - 300M parameters) can outperform much larger and massively multilingual foundation models (with over 1B parameters). This finding is also supported in other works~\cite{haroun2026ara}, where compact domain specialized speech models have proven to be effective.

While these previous works establish important Ewe resources and baselines, the amount of transcribed data remains limited, not exceeding much over 100 hours. Furthermore, one cannot naively combine datasets as projects and datasets can overlap. For example~\cite{diack2026waxal} and \cite{wiafe-etal-2025-building} seem to be part of the same project, and thus we refer to this dataset as WAXAL.

\section{Benchmark Construction}
\label{sec:benchmark}

We consolidate Ewe speech resources with overlapping source collections from the WAXAL project~\cite{diack2026waxal}, BibleTTS~\cite{meyer22c_interspeech}, and the Common Voice - Mina data~\cite{ardila-etal-2020-common} into a single benchmark with predefined training, validation, and test sets. We vary the amount of training data across experiments to create three training splits: one split with training data from WAXAL, a second split with the training data from WAXAL and BibleTTS, and a third split with training data from all three. We keep the validation and test sets unchanged.

We computed a cryptographic hash of the audio content of every file and used it as an identifier, which detects exact duplicates independently of file names, paths, or metadata. We applied this both across corpora and across the training, validation, and test partitions, since concatenating independently released datasets can otherwise inject duplicated recordings into training and evaluation and inflate measured performance. \\

\noindent\textbf{Deduplication:}
After fixing the train, validation, and test splits, we use the audio hash to removed duplicated data. Exact audio-hash comparison identifies 789 recordings shared between training and validation and 699 between training and test, with no audio-hash overlap between validation and test; 52 speakers occur in both training and validation, and 51 in both training and test. 

We further remove any data within each split that contains exact transcript matches after transcript normalization. After this deduplication process, the training, validation, and test partitions are disjoint with respect to both normalized transcripts and exact audio content; the train--validation, train--test, and validation--test transcript intersections are all empty, and no exact audio-hash matches remain across partitions. These checks are designed to prevent identical recordings or normalized transcripts from appearing in both training and evaluation data. Speaker overlap, however, is still possible. We therefore consider the resulting partitions transcript- and exact-audio-disjoint to the extent detectable by our normalization and hashing procedures, but do not claim speaker disjointness. \\

\begin{table}[t]
\centering
\caption{Training, validation, and test set sizes, measured in number of utterances. Validation and test sets are fixed across all configurations to compare performance with the addition of more training data. Data comes from WAXAL~\cite{diack2026waxal}, BibleTTS~\cite{meyer22c_interspeech}, and Common Voice - Mina (MINA)~\cite{ardila-etal-2020-common}.}
\vspace{0.2cm}
\label{tab:datasets}
\begin{tabular}{lrrr}
\toprule   
\textbf{Source} & \textbf{Train} & \textbf{Valid.} & \textbf{Test}\\
\midrule
WAXAL & 16,078 & 1,614 & 1,443 \\
WAXAL + BibleTTS & 36,011 & 1,614 & 1,443 \\
WAXAL + BibleTTS + MINA & 52,417 & 1,614 & 1,443 \\
\bottomrule
\end{tabular}
\end{table}

\noindent\textbf{Dataset statistics:}
Following deduplication, the consolidated dataset contains 52,417 training utterances (158.75~hours), 1,614 validation utterances (8.33~hours), and 1,443 test utterances (7.74~hours), for a total of 174.82~hours of transcribed audio.
To evaluate the contribution of the individual data resources, we define three training configurations of increasing size, with each configuration incrementally adding an additional resource to the preceding one: WAXAL, WAXAL + BibleTTS, and WAXAL + BibleTTS + MINA (Table~\ref{tab:datasets}).
The validation and test sets are held fixed across all configurations.
This incremental setup allows us to examine how performance changes as additional training resources are incorporated, while ensuring that differences between configurations reflect the quantity and composition of the training data rather than changes to the evaluation protocol. \\

\noindent\textbf{Licensing:}
The released package is distributed under CC-BY-4.0, in accordance with the consent, ethical approval, and licensing conditions documented by the upstream WAXAL, BibleTTS, and Common Voice - Mina resources.

\section{Models and Evaluation}
\label{sec:models}

\begin{table}[t]
\centering
\caption{Models and fine-tuning settings.}
\label{tab:hparams}
\setlength{\tabcolsep}{4pt}
\begin{tabular}{@{}lccc@{}}
\toprule
& \multicolumn{2}{c}{Whisper} & \multirow{2}{*}{AfriHuBERT} \\
\cmidrule(lr){2-3}
& medium & base & \\
\midrule
Parameters & 769\,M & 74\,M & 97.4\,M$^{\star}$ \\
Effective epochs & 6.41 & 10.07 & 27 \\
Batch / accum. & 4 / 4 & 4 / 4 & 2 / -- \\
Lang. condition & \texttt{yo} & \texttt{yo} & -- \\
Objective & Seq2Seq & Seq2Seq & CTC \\
\bottomrule
\end{tabular}\\[2pt]

\end{table}

We use two complementary pretrained models to establish baselines: Whisper and AfriHuBERT. Whisper~\cite{pmlr-v202-radford23a} is a multilingual encoder--decoder ASR model trained on approximately 680,000~hours of weakly supervised audio. Although Whisper was not reported to have been trained on Ewe, its training data includes other West African languages. Given this regional language coverage and its strong performance in previous work~\cite{wiafe-etal-2025-building}, we select Whisper as a baseline for fine-tuning. Following prior work on Ewe ASR~\cite{wiafe-etal-2025-building}, we use Yoruba as the language condition for all Whisper experiments. This choice facilitates comparison with their results and reflects their finding that Yoruba conditioning yielded the best performance.

AfriHuBERT~\cite{alabi25_interspeech} is a minimal encoder-only self-supervised model pretrained with the HuBERT framework~\cite{hsu2021hubert}. It was developed by starting with mHuBERT-147~\cite{zanonboito24_interspeech} and continuing the pretraining using more than 10,000~hours of speech covering over 1,200 African languages. We adapt AfriHuBERT to ASR with a CTC objective with the SpeechBrain library~\cite{ravanelli2024open} following the same protocol as other work that has fine-tuned AfriHuBERT for Fongbe (another West African Language)~\cite{kponou25_interspeech}.

We evaluate two Whisper sizes, Whisper-base and Whisper-medium across all versions of the training data. Based on these results, we decided to only fine-tune AfriHuBERT on largest train split, as this led to best performance for the Whisper models. The model and fine-tuning details are summarized in Table~\ref{tab:hparams}. This is not a controlled architecture comparison because the models differ in parameter count, pretraining data, architecture, objective, and compute budget. Instead, they serve as practical baselines evaluated for state-of-the-art models on our consolidated Ewe benchmark.\\

All systems are evaluate on the same 1,443 test utterances. We report word error rate (WER) and character error rate (CER) on the original Ewe orthography, including punctuation and diacritics. 
To assess sensitivity to orthographic convention we additionally report scores after removing punctuation (WER$_{\mathrm{np}}$) and after removing both punctuation and combining diacritical marks (WER$_{\mathrm{np\_nd}}$, CER$_{\mathrm{np\_nd}}$).

\section{Results and Analysis}
\label{sec:results}
Table~\ref{tab:results} reports WER across training splits and
models. As the amount of training data increased so did the model performance. However, the addition of the Common Voice - Mina data did not result in substantial model improvement.

When using training data from all three sources, Whisper-medium obtained the lowest WER at 37.98\%, followed by AfriHuBERT-CTC obtains a higher WER at 39.70\%, and Whisper-base at 44.72\%. The WERs decreases when punctuation is removed, and further when combining diacritical marks are also removed. For Whisper-medium, WER falls from 37.98\% to 33.85\% (np) and 30.95\% (np\_nd); for AfriHuBERT-CTC, from 39.70\% to 35.52\% and 32.80\%. \\

\begin{table}[t]
\centering
\caption{ASR results on the fixed 1,443-utterance test set (\%).
W\,=\,WAXAL, B\,=\,BibleTTS, M\,=\,Common Voice - Mina. 
Columns \emph{np}, \emph{$np\_nd$}, represent performance when punctuation is removed and when punctuation and diacritics are removed, respectively.\\}
\label{tab:results}
\setlength{\tabcolsep}{2pt}
\begin{tabular}{@{}llrrrrr@{}}
\toprule
\multirow{2}{*}{Data} & \multirow{2}{*}{Model} & \multicolumn{3}{c}{WER} & \multicolumn{2}{c}{CER} \\
\cmidrule(lr){3-5}\cmidrule(lr){6-7}
 & & full & np & np\_nd & full & np\_nd \\
\midrule
W & Whisper-base & 46.84 & 42.10 & 39.36 & 16.23 & 14.18 \\
 & Whisper-med & 39.69 & 35.49 & 32.65 & 13.03 & 11.20 \\

\midrule
W+B & Whisper-base & 44.92 & 40.48 & 37.73 & 15.29 & 13.22 \\
 & Whisper-med & 38.71 & 34.32 & 31.35 & 12.48 & 10.50 \\

\midrule
\multirow{3}{*}{W+B+M}
 & Whisper-base & 44.72 & 40.28 & 37.57 & 15.07 & 13.08 \\
 & Whisper-med & \textbf{37.98} & \textbf{33.85} & \textbf{30.95} & \textbf{12.18} & \textbf{10.24} \\
 & AfriHuBERT & 39.70 & 35.52 & 32.80 & 12.76 & 10.91 \\
\bottomrule
\end{tabular}
\end{table}

\begin{table*}[t]
\centering
\caption{Representative errors on the Ewe test set. \emph{Ortho} denotes orthography errors.}
\label{tab:qualitative}
\setlength{\tabcolsep}{2.5pt}
\renewcommand{\arraystretch}{0.95}
\small
\begin{tabular}{@{}p{0.29\linewidth}p{0.29\linewidth}p{0.32\linewidth}@{}}
\toprule
\textbf{Reference} & \textbf{Hypothesis} & \textbf{Error type / analysis} \\
\midrule

& \multicolumn{2}{@{}l}{\textbf{Whisper-medium}}\\

\textit{\dots{}  utsu \textipa{\:d}{eka} le mega foni
 e asi \dots{}} &
\textit{\dots{}  utsu \textipa{\:d}{eka} l\'e megafoni
 e asi \dots{}} &
Word boundary: \textit{mega foni} $\rightarrow$ \textit{megafoni}. \\

\textit{ame geewo tsi atsitre  e goto.} &
\textit{ame geewo tsatsitre e goto.} &
Word boundary: \textit{tsi atsitre} $\rightarrow$ \textit{tsatsitre}. \\

\textit{\dots{} \textopeno i kuku fitaa eye \dots{}} &
\textit{\dots{} \textopeno{} kuku fita eye \dots{}} &
Ortho: \textit{fitaa} $\rightarrow$ \textit{fita}; epenthetic vowel deletion. \\

\textit{\dots{} a\textltailgamma ego violet kple a\textltailgamma ego
 i \dots{}} &
\textit{\dots{} a\textltailgamma ego baulet kple a\textltailgamma ego
 i \dots{}} &
Lexical substitution: \textit{violet} $\rightarrow$ \textit{baulet}. \\

\midrule

& \multicolumn{2}{@{}l}{\textbf{AfriHuBERT-CTC}}\\

\textit{yehowa \textflorin e nya va na solomo be,} &
\textit{yehowa \textflorin e nya va na solo be,} &
Proper-noun: \textit{solomo} $\rightarrow$ \textit{solo}. \\

\textit{ale mebu be ame kukuwo t\textopeno{} nyo s\~a a \dots} &
\textit{ale mebu be ame kukuwwo t\textopeno{} nyo s\~a a \dots} &
Ortho: \textit{kukuwo} $\rightarrow$ \textit{kukuwwo}. \\

\textit{k\textopeno si \textipa{\:d}enyi gava yi. anyigba \textflorin u
az\textipa{O} ml\textipa{O} eba.} &
\textit{k\textopeno si \textipa{\:d}enyi gava yi, anyigba \textflorin u
az\textipa{O} ml\textipa{O} eba.} &
Punctuation: \textit{yi.} $\rightarrow$ \textit{yi,} \\

\textit{yesu bia wo be, ``nu ka dim miele?''} &
\textit{yesu bia wo be, ``nu ka dim miele?} &
Punctuation: closing quotation mark omitted. \\

\bottomrule
\end{tabular}
\end{table*}

\noindent\textbf{Relation to previous Ewe results:}
Earlier Ewe ASR results provide useful context for interpreting our findings, but are not directly comparable because they were obtained using different datasets and experimental settings. Wiafe et al.~\cite{wiafe-etal-2025-building} report WERs of approximately 37\% for Whisper-based fine-tuning on Ewe, while WAXAL-Net~\cite{olufemi2026waxal} reports WERs in the low-to-mid 30\% range when punctuation and capitalization are excluded. These results provide a useful point of reference for the \textit{np} condition in Table~\ref{tab:results}, in which we likewise remove punctuation and capitalization. Under this setting, our Whisper-medium and AfriHuBERT models also achieve WERs in the low-to-mid 30\% range.

The purpose of this comparison is to position our results within the existing Ewe ASR literature rather than to claim a direct, controlled comparison with prior work. Our primary contribution is instead to establish a benchmark under a common evaluation protocol and fixed test subset, enabling consistent comparison among several baselines based on state-of-the-art pretrained models. To support reproducibility, we also provide the code, model configurations, and released artifacts used in our experiments.\\

\noindent\textbf{Qualitative error analysis:} To complement the aggregate CER and WER metrics we inspect reference transcriptions and model hypotheses directly. 
This is important for low-resource languages such as Ewe, where differences in orthography, word segmentation, and closely related lexical forms can incur substantial WER or CER penalties even when the hypothesis stays acoustically and linguistically close to the reference.
Table~\ref{tab:qualitative} gives representative examples of recurring error types on the test set for both models.

For Whisper we note word-boundary and lexical errors.
Word-boundary errors arise when the model merges or separates words written differently from the reference. A small segmentation change can produce several word-level errors even when most of the acoustic content is recognized correctly, resulting in high WERs for simply missing spacing.
Lexical substitutions occur between acoustically similar forms. We found occurrences of these substitutions particularly with loanwords and proper nouns and is most likely a reflection of a limited coverage of these terms training data. These errors tend to be more serious as they can significantly change the meaning of the transcript.

For AfriHuBERT, we note punctuation errors, such as a missing comma or missing close quotation. As AfriHuBERT is a compact encoder-only model that has never seen text, learning proper punctuation on a minimal ASR dataset is challenging. This is in contrast to Whisper, which was trained in a semi-supervised fashion and has potentially seen many examples of punctuation across languages. Similar to Whisper, proper nouns and loanwords also proved to be challenging for AfriHuBERT. However, while Whisper would sometimes perform a lexical substitution, often replacing the whole word with a similarly sounding Ewe word, AfriHuBERT would output something non-Ewe resembling the pronunciation of the proper noun.  This reflects the character level CTC fine-tuning of AfriHuBERT versus Whisper's decoder which operates at the token level. Lastly, both models produced other orthographic errors such as \textit{fitaa} $\rightarrow$ \textit{fita} and \textit{kukuwo} $\rightarrow$ \textit{kukuwwo}.

These cases also show why aggregate WER alone is an incomplete picture: a high-WER utterances may preserve much of the lexical content, while a single substitution of a
linguistically important word carries a small CER/WER penalty, yet changes the intended meaning. 
The examples are illustrative, and thus establishing that a given architecture systematically produces a given error type would require a quantitative analysis of error frequencies over the complete test set.

\section{Conclusion}
\label{sec:conclusion}

We release Nufo, a consolidated Ewe ASR package containing 174.82~hours of speech,
together with a shared evaluation protocol and establish baselines with Whisper and AfriHuBERT. Our fine-tuned models reach WER from around 30\% to 46\% and CER from around 10\% to 16\%, depending on the model and training data.

While we believe this benchmark will facilitate future work in Ewe speech processing, this dataset is still limited. Future work could involve taking advantage of existing unlabeled Ewe speech data with self-supervised learning methods in order to better adapt models to Ewe speech before find-tuning. Expanding the amount and diversity of labeled speech is another important direction, particularly for improving robustness across speakers, dialects, and recording conditions.

\section{Compliance with Ethical Standards}
\label{sec:ethics}

This research study was conducted retrospectively using human subject data
made available in open access by the WAXAL~\cite{diack2026waxal},
BibleTTS~\cite{meyer22c_interspeech}, and Common Voice~\cite{ardila-etal-2020-common}
projects. 
Ethical approval was not required.

\printbibliography

\end{document}